\pdfoutput=1
\documentclass[11pt]{article}
\usepackage[]{acl}
 \usepackage{booktabs}
\usepackage{times}
\usepackage{latexsym}
\usepackage{float}
\usepackage[utf8]{inputenc}
\usepackage[main=english, basque, arabic, french, italian, catalan, spanish, finnish, hindi, swedish, czech]{babel}
\babelprovide[import]{hindi}
\usepackage[T1]{fontenc}
\usepackage{microtype}
\usepackage{inconsolata}
\usepackage{CJKutf8}
\usepackage{graphicx}
\usepackage{tikz}
\usepackage{color,soul}
\usepackage{multirow} 
\usepackage{multicol}
\usepackage{fontawesome5}
\usepackage{xcolor}
\definecolor{gold}{RGB}{248, 187, 87}  
\definecolor{silver}{RGB}{158, 158, 158}  
\definecolor{bronze}{RGB}{206, 115, 50}  

\usepackage{supertabular}
\newcommand*\circled[1]{\tikz[baseline=(char.base)]{
            \node[shape=circle,draw,inner sep=2pt] (char) {#1};}}
\title{AILS-NTUA at SemEval-2025 Task 3: Leveraging Large Language Models and Translation Strategies for Multilingual Hallucination Detection}

\author{
    Dimitra Karkani, Maria Lymperaiou, Giorgos Filandrianos, Nikolaos Spanos, \\  \textbf{Athanasios Voulodimos, Giorgos Stamou} \\ 
    School of Electrical and Computer Engineering,  AILS Laboratory\\
    National Technical University of Athens \\
    \texttt{\href{mailto:dkarkanh@gmail.com}{dkarkanh@gmail.com}, \{\href{mailto:marialymp@islab.ntua.gr}{marialymp}, \href{mailto:geofila@islab.ntua.gr}{geofila}, \href{mailto:nspanos@ails.ece.ntua.gr}{nspanos}\}@ails.ece.ntua.gr, } \\
    \texttt{\href{mailto:thanosv@mail.ntua.gr}{thanosv@mail.ntua.gr},
    \href{mailto:gstam@cs.ntua.gr}{gstam@cs.ntua.gr}}\\
}

\begin{document}
\maketitle
\begin{abstract}
Multilingual hallucination detection stands as an underexplored challenge, which the Mu-SHROOM shared task seeks to address. In this work, we propose an efficient, training-free LLM prompting strategy that enhances detection by translating multilingual text spans into English. Our approach achieves competitive rankings across multiple languages, securing two first positions in low-resource languages. The consistency of our results highlights the effectiveness of our translation strategy for hallucination detection, demonstrating its applicability regardless of the source language.
\end{abstract}

\section{Introduction}
Hallucinations in Large Language Models (LLMs) pose a significant challenge, as they can generate fluent yet factually incorrect or misleading content 
\cite{zhang2023sirenssongaiocean, Huang_2025}. Several detection techniques have been developed, either by accessing the LLM's internals \cite{azaria-mitchell-2023-internal, sriramanan2024llmcheck, Chen2024INSIDELI}, probing generation inconsistencies \cite{manakul-etal-2023-selfcheckgpt, Mndler2023SelfcontradictoryHO} or exploiting token-based classification \cite{Quevedo2024DetectingHI}. 

While hallucination detection has been widely studied in monolingual contexts, multilingual settings are severely underexplored and accompanied by additional complexities. Variations in linguistic structure, resource availability, and training data distribution can lead to uneven model reliability across languages. Low-resource languages, in particular, are more susceptible to hallucinations due to limited high-quality training data, making factual consistency a critical issue. Recent works in multilingual hallucinations detection verify faithfulness shortcomings \cite{qiu-etal-2023-detecting, shen-etal-2024-multilingual}, shift the focus on data rather than model capacity \cite{Guerreiro2023HallucinationsIL}
while also pinpointing evaluation concerns \cite{kang2024comparinghallucinationdetectionmetrics}.

To this end, the Mu-SHROOM shared task is proposed in order to fill this gap by providing high-quality data on 14 languages, including low-resource languages such as Farsi, Czech, Finnish, Swedish and Basque. This effort is accompanied by annotations on hallucinated data spans within given sentences, which participants have to automatically detect. 

In our work, we leverage LLM prompting and translation strategies to address hallucination detection without designing independent systems per language. Particularly, we combine two LLMs, Llama 3 \cite{grattafiori2024llama3herdmodels} and Claude \cite{TheC3}, prompting them in a few-shot manner to detect hallucinatory spans. 
Our language-adaptive system is proven efficient and successful in both high- and low-resource languages \textit{without any training or fine-tuning}. As a result, we achieve first position in the low-resource Farsi and Czech languages, second position in the high-resource Italian language and among the top 15\% positions in English, Spanish, German, Hindi and Basque.

\section{Background}
\subsection{Task description} Mu-SHROOM is a multilingual extension of SHROOM \cite{shroom-2024-semeval} comprising 14 languages: Arabic (Modern standard)-\textit{AR}, Basque-\textit{EU}, Catalan-\textit{CA}, Chinese (Mandarin)-\textit{ZH}, Czech-\textit{CS}, English-\textit{EN}, Farsi-\textit{FA}, Finnish-\textit{FI}, French-\textit{FR}, German-\textit{DE}, Hindi-\textit{HI}, Italian-\textit{IT}, Spanish-\textit{ES}, and Swedish-\textit{SV}. Participants are tasked to detect hallucinatory spans within generated text as accurately as possible, so that if the span was omitted the hallucination would be removed.

\subsection{Related work}
\textbf{Multilingual NLP Hallucinations}. While hallucination detection has been actively researched in monolingual contexts across various fields \cite{dhuliawala2023chainofverificationreduceshallucinationlarge, manakul2023selfcheckgptzeroresourceblackboxhallucination, min2023factscorefinegrainedatomicevaluation, fabbri-etal-2022-qafacteval, maynez-etal-2020-faithfulness, scialom-etal-2021-questeval}, its multi-lingual counterpart has primarily focused on identifying hallucinations in machine translation, where such hallucinations are defined as translations containing information completely unrelated to the input \cite{10.1162/tacl_a_00615}. Machine translation also has established benchmarks \cite{dale2023halomimanuallyannotatedbenchmark}, as well as metrics \cite{kang2024comparinghallucinationdetectionmetrics}, for evaluating model performance in cross-lingual generation and transfer. Multilingual hallucinations are also extensively studied in text summarization applications. In low-resource languages, summaries are often translated into high-resource languages, such as English, to utilize more reliable evaluation metrics \cite{qiu2023detectingmitigatinghallucinationsmultilingual}. Mu-SHROOM, based on last year's SHROOM challenge \cite{shroom-2024-semeval}, is the inaugural benchmark for multilingual hallucination detection, addressing a significant gap in low-resource language research due to the absence of established benchmarks.
\section{System Overview}
\label{sec:overview}
We focus on LLM prompting and translation strategies to tackle hallucination detection challenges in a language-agnostic manner. Due to the prompt-heavy nature of our approach, no further training is required to attain high scores in either high- or low-resource languages. 
Specifically, we combine two LLMs, Llama 3.1 405B\footnote{meta.llama3-1-405b-instruct-v1:0} and Claude 3.5 Sonnet\footnote{anthropic.claude-3-5-sonnet-20241022-v2:0}, prompting them in a few-shot (FS) way to detect hallucination spans. To improve detection performance in low-resource languages, we also experiment with incorporating a translation tool to translate original input-output data to English. Finally, given the inputs of each MuSHROOM instance, we instruct Llama and Claude to generate the corresponding output and incorporate it as a \textit{hypothesis} to facilitate hallucination detection.

Based on the results of our preliminary experiments presented in App.\ref{sec:preliminaries}, our final system consists of three components, i.e. three experiments:

\textbf{Component 1} We prompt Claude to detect hallucination spans given the input text, the output text in the original language and their translations in English, as well as the outputs produced by Llama as hypothesis.

\textbf{Component 2} We prompt Llama to detect hallucination spans given the input text, the output text in the original language and their translations in English, as well as the outputs produced by Claude as hypothesis.

\textbf{Component 3} We prompt Llama to detect hallucination spans given the input text, the output text in the original language and their translations in English \textit{without} providing extra generated answers as hypothesis. 
We adopt this approach since generated hypotheses can themselves contain hallucinations, resulting in misleading outcomes. Moreover, the LLMs sometimes place undue emphasis on the provided generated hypothesis rather than relying on their internal knowledge, causing
them to miss hallucinatory spans in the outputs.

Each component produces a list of hallucination spans and then the three lists are combined as follows: for each produced span, the assigned probability is calculated as the ratio of the experiments that characterize it as hallucination over the total number of experiments (three).

\section{Methods}
\label{sec:methods}
\paragraph  {Hallucination categorization} To initiate the hallucination detection process, we begin by defining what constitutes a hallucination. This initial categorization allows us to effectively handle data from various tasks and domains. Drawing on \citet{Huang_2025} and our exploratory analysis of both the task's sample and validation data, we identify four distinct types of hallucinations:

\circled{1} \textbf{Input-Output inconsistency}: The produced output is inconsistent with the input, i.e. it does not satisfy the input query or is irrelevant to it.

\circled{2} \textbf{Factual inconsistency}: The output contains information that is factually inconsistent  in a sense that it cannot be associated with verifiable real-world facts.

 \circled{3}  \textbf{Internal output inconsistency}:  The output contains contradictory facts, i.e.c---c- in the generated text span there is inconsistent information. 
 
 \circled{4} \textbf{Misspellings}: The output contains misspelled words.

\paragraph{Output Format}
After defining hallucinations, we specify the expected output format to effectively process LLM outputs. For that purpose we attempt two approaches: In our first approach, in order to make the procedure simpler, we split the output text in parts and prompt the LLMs given the input, the output and a specific part of the output to decide whether that specific part contains a hallucination. This technique is not successful as shown in Tables \ref{tab:iou_shots}, \ref{tab:cor_shots} in the preliminary column. In the second approach, which we ultimately adopt, we prompt the LLM with both the input and output to detect hallucination spans while also encouraging chain-of-thought (CoT) reasoning.

\paragraph{Prompting strategies}
After initializing the process of hallucination detection by providing the hallucination definition and the expected output format, we experiment in both a \textbf{zero-shot (ZS)} and a \textbf{few-shot(FS)} way. In the FS scenario we present an example for each aforementioned category, together with the expected output format. 
For the main process, we adopt the system/user prompt. After consolidating the
system prompts, we experiment with the user prompts and specifically the data given as input to the LLMs. For the simplest approach, we just provide the input-output pair to the LLM. Then, we attempt to enhance performance by also demonstrating the \textit{translations} of inputs and outputs. To deploy our final system, as explained above, we also supply a \textit{hypothesis} generated by the LLMs. To show how each of the steps of the procedure we propose improves the final results we present the following experiments:

\textbf{1. Standalone prompting} We prompt Claude or Llama at a time in either a ZS or a FS manner without translating in English or leveraging  hypotheses.

\textbf{2. Prompting + Translation} We prompt Llama and Claude to detect hallucination spans using the input and output texts in the original language, as well as their English translations, without providing additional generated hypotheses; then, we combine the two lists of produced hallucination spans.

\textbf{3. Prompting + Translation + Hypothesis} 
We prompt Claude to detect hallucination spans using the input and output text in the original language, their English translations, and the outputs produced by Llama as hypotheses. Conversely, we prompt Llama with the same inputs but use Claude's outputs as hypotheses. Additionally, we incorporate the results from Llama’s previous experiment, combining the three produced hallucination span lists.

\paragraph{Translation}


Given the disparity in linguistic resources across different languages and our objective of developing a system that performs robustly across multilingual settings, we investigate various strategies to address this challenge. Specifically, we examine the following key questions in the context of multilingual hallucination detection: \textit{``Is it more effective to provide both the input and output in their original languages and allow the LLM to detect hallucinations, or should we instead supply their English translations?''} Furthermore, \textit{``If input-output pairs are provided in their original language, should the prompt also be in the same language, or is it preferable to present it in English?''}. Conversely, \textit{``If we supplement the detection process with translated input-output pairs, is it more effective for the LLM itself to handle the translation before the detection step, or should an external translation system be employed?''} 

To explore these questions, we conduct the following experiments. Firstly, we experimented with the impact of the language of the prompt given the input-output pairs in their original languages. In this direction, the following \textit{\textbf{Original Input-Output Pairs}} experiments are conducted.

\textbf{No translator}
The simplest approach is to provide the definition of hallucination, along with the examples per category and instructions for the output format in English, and then present the input-output pairs in their original language.

\textbf{External translator - original language}
Given the input-output pairs in their original language, we translate the prompts—which include the hallucination definition and output format instructions—into the original language. To achieve this, we use the Google Translate API for Python\footnote{\url{https://pypi.org/project/googletrans/}}.

For the second part of our experiments, we translate  the input-output pairs into English, the highest-resource language, and then use the translated pairs to detect hallucinations, while the prompt remains in English. The LLM is exposed to both the original language (before translation), as well as with its English version to ensure fairness.
The experiments we conduct belong to the \textit{\textbf{Translated Input-Output Pairs}} category.

\textbf{External translator - English}
We translate the input-output pairs into English using Google Translate and then prompt the LLMs (providing \textit{both} the original and English versions of data) to generate a CoT for hallucination detection in English. 

\textbf{LLM as the translator}
We prompt the LLMs to perform the analysis in two steps: Firstly to translate input-output pairs into English when the text is in another language, and then based on that to detect  hallucinations in the same chat, thus ensuring that the LLM is exposed in both languages before concluding to the hallucination spans identified.

\begin{table*}[ht!]
    \centering \small
    \renewcommand{\arraystretch}{0.6}      
    \setlength{\tabcolsep}{4pt}           
    \begin{tabular}{l|c|c|c|c|c|c}
    \hline
    \textbf{Language (id)} 
    & \textbf{Baseline}
      & \textbf{Preliminary} 
      & \textbf{ZS} 
      & \textbf{FS} 
      & \textbf{FS + Translation}
      & \textbf{FS + Translation + Hypothesis} \\
    \hline
    
    Arabic (ar)& 0.04/0.36/0.05
      & 0.223
      & 0.379
      & 0.425
      & 0.527
      & \textbf{0.584} \\
    
    Catalan (ca)& 0.05/0.24/0.08
      & 0.273
      & 0.482
      & 0.540
      & 0.675
      & \textbf{0.703} \\
    
    Czech (cs)& 0.10/0.26/0.13
      & 0.301
      & 0.388
      & 0.448
      & 0.556
      & \textbf{0.587} \\
    
    German (de)& 0.03/0.35/0.03
      & 0.199
      & 0.531
      & 0.564
      & 0.578
      & \textbf{0.587} \\
    
    English (en)& 0.03/0.35/0.03
      & 0.223
      & 0.425
      & 0.487
      & -
      & \textbf{0.555} \\
    
    Spanish (es)& 0.07/0.19/0.09
      & 0.239
      & 0.385
      & 0.454
      & 0.468
      & \textbf{0.500} \\
    
    Basque (eu)& 0.02/0.37/0.01
      & 0.299
      & 0.431
      & 0.458
      & 0.518
      & \textbf{0.571} \\
    
    Farsi (fa)& 0.00/0.20/0.00
      & 0.202
      & 0.492
      & 0.558
      & 0.687
      & \textbf{0.753} \\
    
    Finnish (fi)& 0.01/0.49/0.00
      & 0.210
      & 0.464
      & 0.529
      & 0.635
      & \textbf{0.683} \\
    
    French (fr)& 0.00/0.45/0.00
      & 0.251
      & 0.447
      & 0.499
      & 0.535
      & \textbf{0.617} \\
    
    Hindi (hi)& 0.00/0.27/0.00
      & 0.189
      & 0.581
      & 0.624
      & 0.709
      & \textbf{0.726} \\
    
    Italian (it)& 0.01/0.28/0.00
      & 0.267
      & 0.597
      & 0.657
      & 0.774
      & \textbf{0.802} \\
    
    Swedish (sv)& 0.03/0.53/0.02
      & 0.276
      & 0.492
      & 0.537
      & 0.585
      & \textbf{0.601} \\
    
    Chinese (zh)& 0.02/0.47/0.02
      & 0.200
      & 0.212
      & 0.304
      & 0.378
      & \textbf{0.419} \\
    \hline
    \end{tabular}
        \caption{Prompting scenarios comparison -- IoU metric. The three baselines are: neural/ mark-all/ mark-none. The best-performing method per language is in \textbf{bold}. This Table considers the best translation strategy.}
    \label{tab:iou_shots}
\end{table*}

\begin{table*}[ht!]
    \centering \small
    \renewcommand{\arraystretch}{0.6}
    \setlength{\tabcolsep}{4pt}
    
    \begin{tabular}{l|c|c|c|c|c|c}
    \hline
    \textbf{Language (id)} 
    & \textbf{Baseline}
      & \textbf{Preliminary} 
      & \textbf{ZS} 
      & \textbf{FS} 
      & \textbf{FS + Translation}
      & \textbf{FS + Translation+hypothesis} \\
    \hline
    
    Arabic (ar)& 0.11/0.01/0.01
      & 0.190
      & 0.484
      & 0.636
      & 0.601
      & \textbf{0.612} \\
    
    Catalan (ca)& 0.06/0.06/0.06
      & 0.397
      & 0.610
      & 0.564
      & 0.700
      & \textbf{0.709} \\
    
    Czech (cs)& 0.05/0.10/0.10
      & 0.368
      & 0.480
      & 0.419
      & 0.557
      & \textbf{0.590} \\
    
    German (de)& 0.11/0.01/0.01
      & 0.333
      & 0.583
      & 0.466
      & 0.614
      & \textbf{0.629} \\
    
    English (en)& 0.11/0.00/0.00
      & 0.357
      & 0.511
      & \textbf{0.635}
      & -
      & 0.628 \\
    
    Spanish (es)& 0.04/0.01/0.01
      & 0.456
      & 0.464
      & 0.547
      & 0.537
      & \textbf{0.565} \\
    
    Basque (eu)& 0.10/0.00/0.00
      & 0.401
      & 0.530
      & 0.555
      & 0.524
      & \textbf{0.566} \\
    
    Farsi (fa)& 0.11/0.01/0.01
    & 0.378
      & 0.583
      & 0.547
      & 0.684
      & \textbf{0.737} \\
    
    Finnish (fi)& 0.09/0.00/0.00
      & 0.478
      & 0.552
      & 0.584
      & \textbf{0.666}
      & 0.652 \\
    
    French (fr)& 0.02/0.00/0.00
      & 0.254
      & 0.564
      & \textbf{0.617}
      & 0.609
      & 0.614 \\
    
    Hindi (hi)& 0.14/0.00/0.00
      & 0.565
      & 0.666
      & 0.676
      & 0.754
      & \textbf{0.760} \\
    
    Italian (it)& 0.08/0.00/0.00
      & 0.526
      & 0.692
      & 0.765
      & 0.757
      & \textbf{0.817} \\
    
    Swedish (sv)& 0.10/0.01/0.01
      & 0.194
      & 0.502
      & 0.525
      & 0.535
      & \textbf{0.562} \\
    
    Chinese (zh)& 0.08/0.00/0.00
      & 0.264
      & 0.317
      & 0.401
      & \textbf{0.487}
      & 0.464 \\
    \hline
    \end{tabular}
    \caption{Prompting scenario comparison -- Correlation. The three baselines are: neural/ mark-all/ mark-none.The best-performing method per language is in \textbf{bold}. This Table considers the best translation strategy.}
    \label{tab:cor_shots}
\end{table*}

\section{Experimental setup}
\paragraph{Dataset}
For the results presented, the test set provided by the task organizers is used. The test set is presented in detail in Appendix \ref{sec:exploratory}.
\paragraph{Baselines} presented by the organizers comprise a neural-based model, and the edge cases of mark-all and a mark-none.
\paragraph{Evaluation} comprises two character-level metrics: first, Intersection-over-Union \textbf{(IoU)} of characters marked as hallucinations in the gold reference vs. characters predicted as such; second, the \textbf{correlation} between the hallucination probabilities occurring from the detection system  and the gold reference probabilities provided by the annotators.

\paragraph{Computational resources} All our experiments are executed in Amazon Bedrock using Google Colab platform for the API calls.

\section{Results}
The results of our experiments are shown in detail in Tables \ref{tab:iou_shots}, \ref{tab:cor_shots} for prompting experiments (IoU and correlation metrics respectively) and in Table \ref{tab:translation_comparison_cor} regarding translation experiments.
In the prompting experiments, the FS approach for both hallucination definition and expected output format significantly improves the results:
The detected hallucination spans are more accurate and the output format is strictly followed, which is fundamental in order to automatically handle the answers and extract the feedback provided by the LLMs. Furthermore, incorporating the English translation of the texts appears to enhance the LLM's performance. A similar effect is observed when integrating the hypothesis from the other LLM. In this case, the LLM is able to compare the hypothesis with the actual output provided to  determine more effectively the presence of hallucinations and identify their respective spans. Notably, these patterns remain consistent across all languages, \textit{regardless of whether they are low-resource or high-resource}. However, the addition of translations and hypotheses has \textit{a more pronounced impact on low-resource} languages compared to high-resource ones.
\begin{table*}[ht!]
    \centering \small
    \renewcommand{\arraystretch}{0.5}
    \setlength{\tabcolsep}{4pt}
       \begin{tabular}{l|c|c|c|c}
        \hline

       \multirow{2}{*}{\textbf{Language (id)}} & \multicolumn{2}{c|}{\textbf{Original Input-Output Pairs}} & \multicolumn{2}{c}{\textbf{Translated Input-Output Pairs}} \\ 
       
        & \textbf{No Translation} & \textbf{External transl. - Original} & \textbf{LLM Translator} & \textbf{External Transl. English} \\ \hline
        Arabic (ar)   & 0.47/0.55 & 0.32/0.40 & 0.61/0.51 & \textbf{0.58/0.61} \\
        
        Catalan (ca)  & 0.46/0.58 & 0.50/0.62 & 0.49/0.59 & \textbf{0.70/0.71} \\
        
        Czech (cs)    & 0.39/0.42 & 0.37/0.43 & 0.42/0.43 & \hspace{-14pt} {\color{silver}\faTrophy} \textbf{0.59/0.59} \\
        
        German (de)   & 0.50/0.51 & 0.47/0.56 & 0.48/0.54 & \hspace{-14pt} {\color{silver}\faTrophy} \textbf{0.59/0.63} \\
        
        English (en)  & \hspace{-14pt} {\color{bronze}\faTrophy} \textbf{0.55/0.63} & - & -     & -     \\
        
        Spanish (es)  & 0.49/0.49 & 0.31/0.477 & 0.42/0.480 & \hspace{-14pt} {\color{silver}\faTrophy} \textbf{0.50/0.56} \\
        
        Basque (eu)   & 0.35/0.46 & 0.34/0.44 & 0.37/0.49 & \textbf{0.57/0.57} \\
        
        Farsi (fa)    & 0.50/0.61 & 0.49/0.58 & 0.52/0.63 & \hspace{-23pt} {\color{silver}\faTrophy} \textbf{0.75/0.74} \\
        
        Finnish (fi)  & 0.54/0.57 & 0.54/0.56     & 0.53/0.58 & \hspace{-23pt} {\color{bronze}\faTrophy} \textbf{0.68/0.65} \\
        
        French (fr)   & 0.49/0.530 & 0.43/0.450 & 0.45/0.46 & \hspace{-14pt} {\color{silver}\faTrophy}    \textbf{0.617/0.614} \\
        
        Hindi (hi)    & 0.65/0.67 & 0.66/0.68 & 0.70/0.710 & \hspace{-18pt} {\color{bronze}\faTrophy} \textbf{0.73/0.760} \\
        
        Italian (it)  & 0.62/0.620 & 0.604/0.679 & 0.730/0.680 & \hspace{-14pt} {\color{gold}\faTrophy}    \textbf{0.802/0.817}  \\
        
        Swedish (sv)  & 0.53/0.550 & 0.555/0.567    & 0.570/0.540 & \hspace{-14pt} {\color{gold}\faTrophy} \textbf{0.601/0.562} \\
        
        Chinese (zh)  & 0.343/0.399 & 0.399/0.388 & 0.379/0.333 & \hspace{-14pt} {\color{silver}\faTrophy} \textbf{0.419/0.464} \\
        \hline
    \end{tabular}
        \caption{Translation performance comparison - IoU/Correlation metrics respectively. The best-performing method per language is in \textbf{bold}. The best translation strategy is used in the results presented in Tables \ref{tab:iou_shots}, \ref{tab:cor_shots}.}
    \label{tab:translation_comparison_cor}
\end{table*}

\begin{table*}[h!]
\centering\small
\begin{tabular}{p{0.3cm}|p{14.7cm}}
\hline
\textbf{id}& \textbf{Sentence} \\
\hline
ca &  El municipi de Yushu es troba a \colorbox{pink}{4.500} metres sobre el nivell del mar.\\
cs  & Řeka Labe (německy Elbe) pramení v \colorbox{pink}{Českém lese}, konkrétně v okrese \colorbox{pink}{Jičín}, v nadmořské výšce \colorbox{pink}{816} metrů. Pramení v \colorbox{pink}{údolí mezi vrcholy Kozákov} (\colorbox{pink}{744} metrů) \colorbox{pink}{a Říp} (\colorbox{pink}{459} metrů). \\
de & Mario Bola\colorbox{pink}{ti} wechselt im Jahr \colorbox{pink}{1998} zum Verein \colorbox{pink}{AC Mailand}. \\
en & Mouthier is located in the department of \colorbox{pink}{Haute-Loire}. \\
eu & \colorbox{pink}{Hiru} espezie bakarrik daude. \\
fi & Folorunsho Alakija on nigerialainen \colorbox{pink}{kirjailija} ja \colorbox{pink}{aktivisti}. Hän on kirjoittanut useita \colorbox{pink}{kirjoja, muun muassa "The Slave Girl" ja "The Slave Girl's Daughter", jotka käsittelevät naisten sortoa ja orjuutta}. \\
fr & L'espèce Pseudomugil gertrudae appartient à la famille des \colorbox{pink}{Poeciliidae}, qui est une famille d'espèces de poissons \colorbox{pink}{d'eau douce et d'eau salée. Elle est également connue sous le nom de poisson-chat de Gertrude ou de } \\
& \colorbox{pink}{poisson-chat de Gertrude}.
\\
it & Il produttore dell'album "Plastic Letters" di Blondie fu \colorbox{pink}{Mike Chapman}.
\\
sv & David Sandbergs födelseort är \colorbox{pink}{New York}. \\
zh & \begin{CJK}{UTF8}{gbsn}新缬草\colorbox{pink}{原产于欧洲，特别是地中海沿岸地区，包括西班牙、葡萄牙、法国南部、意大利和希腊等地}\end{CJK}\\
& \begin{CJK}{UTF8}{gbsn}。\colorbox{pink}{它在这些地区的自然环境中广泛分布，并且在园艺上也被引种到其他地区。由于其}\end{CJK} \\
& \begin{CJK}{UTF8}{gbsn}\colorbox{pink}{美丽的花朵和耐旱的特性，新 Valerie 在全球各地都有一定的栽培和观赏价值}\end{CJK} \\
\hline
    \end{tabular}
    \caption{Qualitative results in various languages. The detected hallucination span is \colorbox{pink}{highlighted}.}
    \label{tab:qual-main}
\end{table*}

Regarding the translation experiments, interesting insights emerge. On one hand, in the simplest approach—where prompts are given in English while pairs remain in their original language—the English language score is not the highest. This suggests that translation aids in hallucination detection and partially addresses the challenges of low-resource languages. However, the reverse process, translating into the language of the pairs, does not appear to offer the same benefits.
Additionally, the use of the Google Translator is more effective compared to the end-to-end system where the LLMs are prompted to translate the input and output texts themselves. Thus, the most effective approach for identifying multilingual hallucinations is to provide \textit{both the prompt and the input-output pairs in English}, using an \textit{external translation system} rather than incorporating translation as a step within the LLM pipeline. This finding holds consistently across all languages in the dataset.

The results tables also show that for high-resource languages such as Spanish, Chinese, and German, the FS scenario and the incorporation of the generated hypothesis contribute the most towards performance improvements. In contrast, \textit{for low-resource languages, translation is a crucial component} in achieving similar results. The prompts are detailed in App. \ref{sec:prompts}.

As a demonstration of our system, in Table \ref{tab:qual-main} we present some qualitative results from our system which achieve a perfect score of IoU=1.

\section{Conclusion}
In this work, we detect multilingual hallucination spans in the outputs from the SemEval 2025-Task 3 MuSHROOM dataset using LLM prompting and translation techniques. We showcase the merits of converting all data in English, especially for low-resource languages, achieving highly-ranked results in a language-agnostic manner overall.
\bibliography{main}

\appendix

\section{Exploratory data analysis}
\label{sec:exploratory}
\subsection{Sample set}
In the preliminary phase of the competition, a small sample set is released to allow participants to acclimate to the task.
The sample set consists of a total of 8 samples in 3 different languages (3 samples in English, 3 samples in Spanish and 2 samples in French). 
The features of each samples are:
\begin{itemize}
    \item 'id': a unique number of the sample.
    \item 'lang': the id of the language of the participating text.
    \item 'model\_input': the input prompt given to the model.
    \item 'model\_output\_text: the output text the model generated.
    \item 'model\_id ': the id of the model that produced the output.
    \item 'soft\_labels':  spans that include a start and end character number, together with an assigned probability to the span constrained by these two characters.
    \item 'hard\_labels': spans for which the assigned soft label probability is more than 0.5.
\end{itemize}

Models producing hallucinations to be detected are the following:
The different models used are:
\begin{itemize}
    \item Model id: TheBloke/Mistral-7B-Instruct-v0.2-GGUF (4 annotated samples).
    \item Model id:meta-llama/Meta-Llama-3-8B-Instruct (1 annotated sample).
    \item Model id:Iker/Llama-3-Instruct-Neurona-8b-v2 (3 annotated samples).
\end{itemize}

Since we propose a model-agnostic approach, information regarding the model from which hallucination occurs is overlooked in practice.

\subsection{Validation set}
The validation set consists of 10 subsets of 50 samples each, separated by language, comprising 500 samples in total. The different languages are: arabic, german, english, spanish, finnish, french, hindi, italian, swedish and chinese. 
The features of the samples were the same as the ones of the sample set in version 1, with the extension of:
\begin{itemize}
    \item model\_output\_tokens: the tokens of the output of the model
    \item model\_output\_logits: the logits of the output of the model
\end{itemize}
in version 2 that was released after the evaluation phase.

Since the data were of different tasks and domains, the exploratory analysis contained an effort to categorize the hallucinations found in the data. 

Based on the definition of hallucinations and overgeneration mistakes, we distinguish the following types of hallucinations:

A hallucination is the production of fluent but incorrect output of an LLM.  The definition of incorrect output falls in four categories:

1. The \textit{output is inconsistent with the input}, so the produced answer does not answer the input query or is irrelevant to it.

2. The \textit{output contains a factual inconsistency}, so contains something that is not a validated fact or is wrong.

3. The \textit{output contains contradictory facts}, so in the output there are things that cannot be true at the same time.

4. The \textit{output contains misspelled words}.

Based on these categories, in an attempt to understand better the annotation procedure, we conducted manually a categorization of the hallucinations marked in the outputs:
\begin{itemize}
\item Input-Output Inconsistency: 10
\item Factual Inconsistency: 31
\item Output conflicts: 6
\item Misspelled words: 8
\end{itemize}

\subsection{Test set} 
\begin{figure}[h!]
    \centering
    \includegraphics[width=\linewidth]{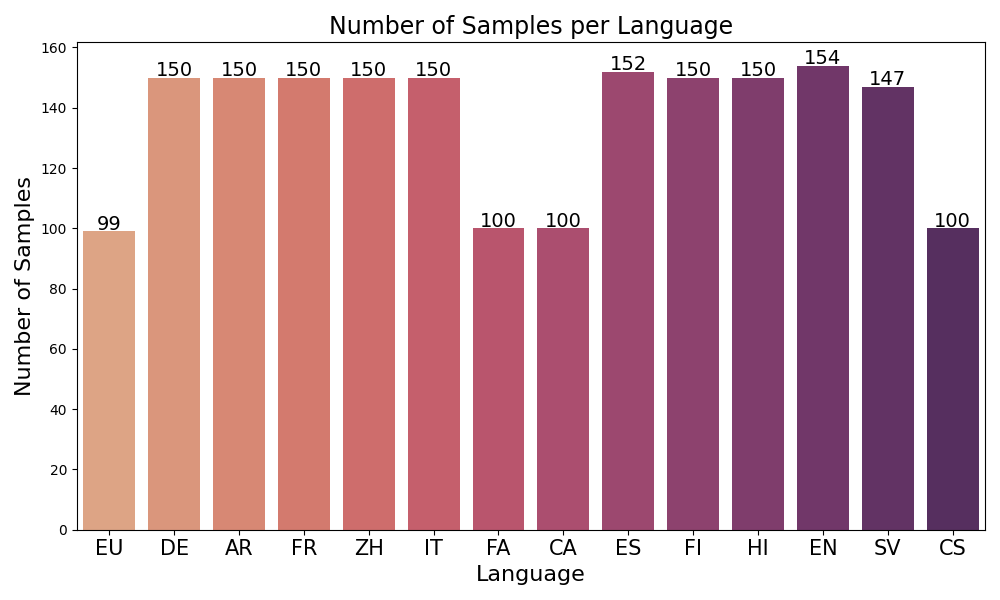}
    \caption{Number of annotated test data samples per language.}
    \label{fig:samples-per-language}
\end{figure}

The number of annotated samples per language in the test set are presented in Figure \ref{fig:samples-per-language}. Most languages contain around 150 samples, with the exception of Basque (EU), Farsi (FA), Catalan (CA) and Czech (CS) that contain around 100 samples each.

\paragraph{Input \& Output sentence length}

In the following Figures, we present the length distribution regarding the input and output sentences per language on the test set.

\begin{figure}[h!]
    \centering
    \includegraphics[width=0.99\linewidth]{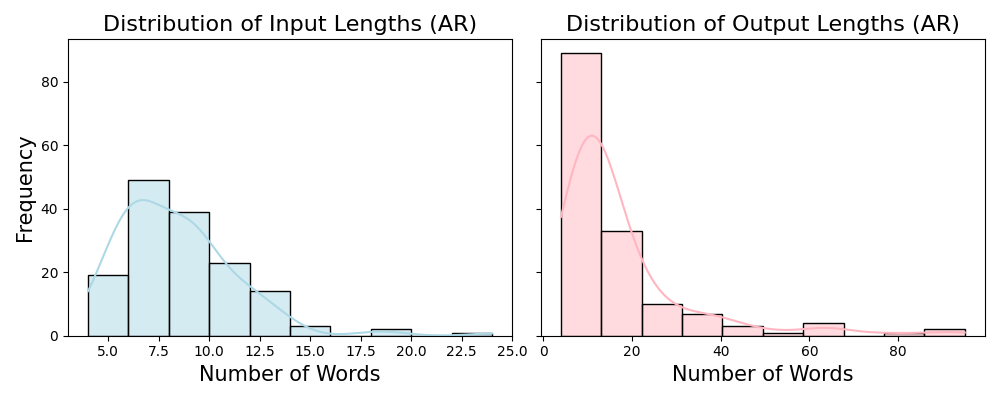}
    \caption{Input/Output length distribution for Arabic.}
    \label{fig:enter-label}
\end{figure}

\begin{figure}[h!]
    \centering
    \includegraphics[width=0.99\linewidth]{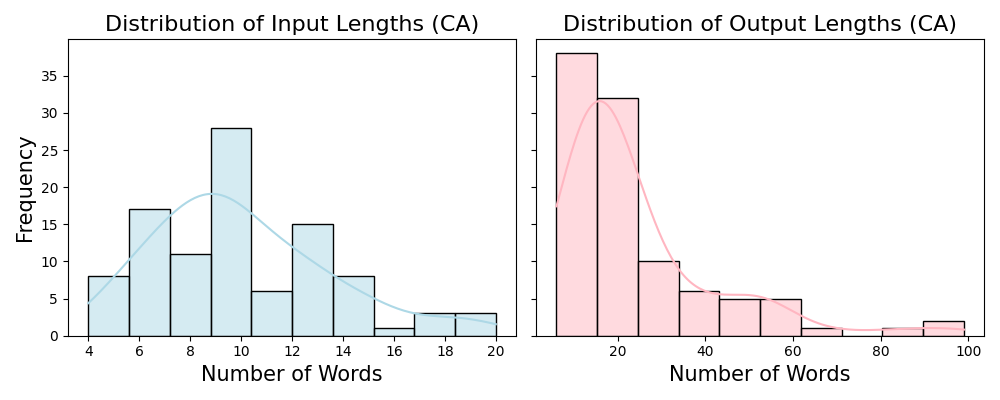}
    \caption{Input/Output length distribution for Catalan.}
    \label{fig:enter-label}
\end{figure}

\begin{figure}[h!]
    \centering
    \includegraphics[width=0.99\linewidth]{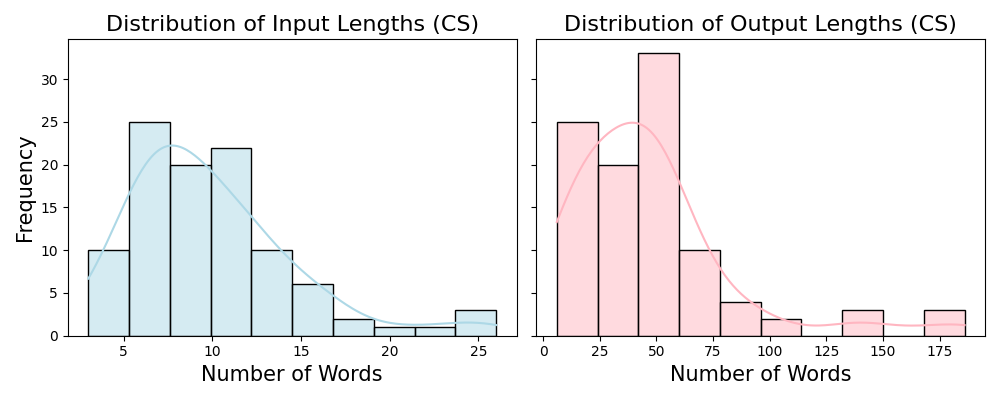}
    \caption{Input/Output length distribution for Czech.}
    \label{fig:enter-label}
\end{figure}

\begin{figure}[h!]
    \centering
    \includegraphics[width=0.99\linewidth]{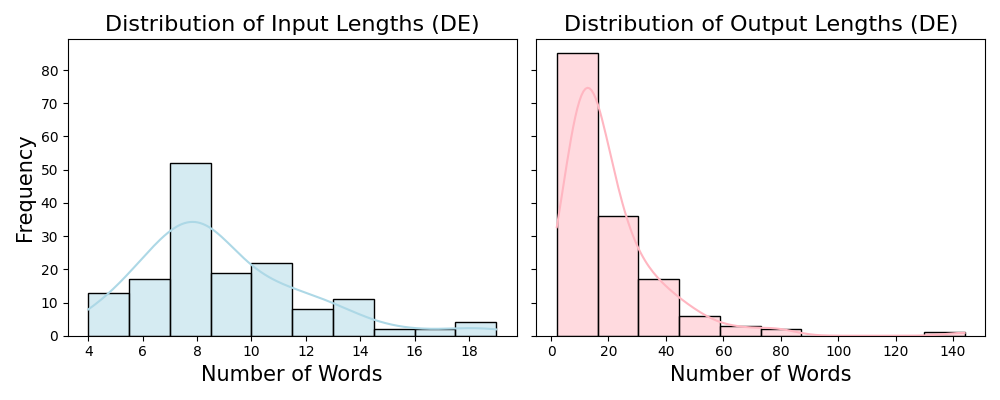}
    \caption{Input/Output length distribution for German.}
    \label{fig:enter-label}
\end{figure}

\begin{figure}[h!]
    \centering
    \includegraphics[width=0.99\linewidth]{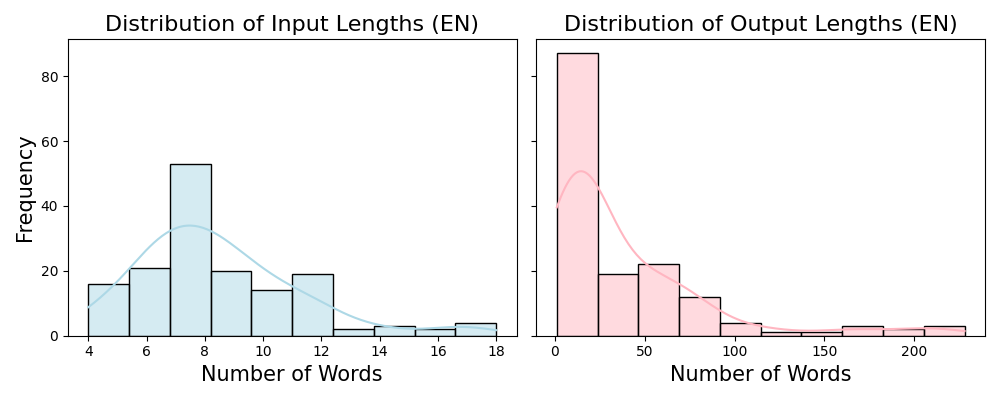}
    \caption{Input/Output length distribution for English.}
    \label{fig:enter-label}
\end{figure}

\begin{figure}[h!]
    \centering
    \includegraphics[width=0.99\linewidth]{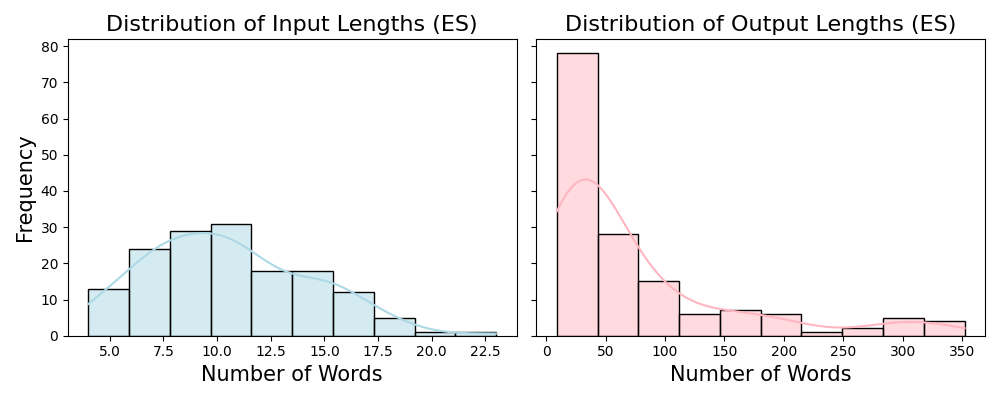}
    \caption{Input/Output length distribution for Spanish.}
    \label{fig:enter-label}
\end{figure}

\begin{figure}[h!]
    \centering
    \includegraphics[width=0.99\linewidth]{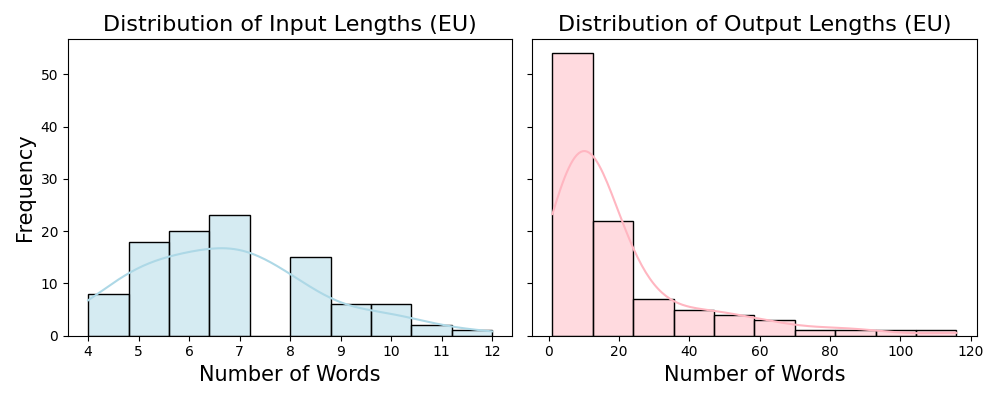}
    \caption{Input/Output length distribution for Basque.}
    \label{fig:enter-label}
\end{figure}

\begin{figure}[h!]
    \centering
    \includegraphics[width=0.99\linewidth]{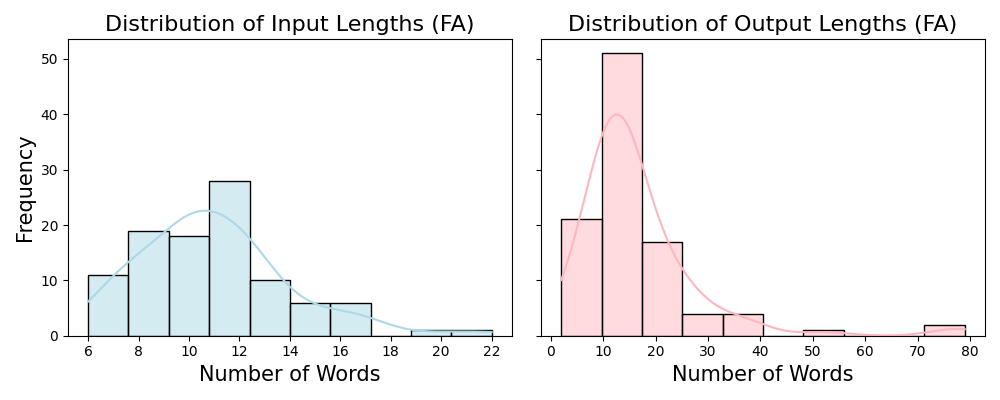}
    \caption{Input/Output length distribution for Farsi.}
    \label{fig:enter-label}
\end{figure}

\begin{figure}[h!]
    \centering
    \includegraphics[width=0.99\linewidth]{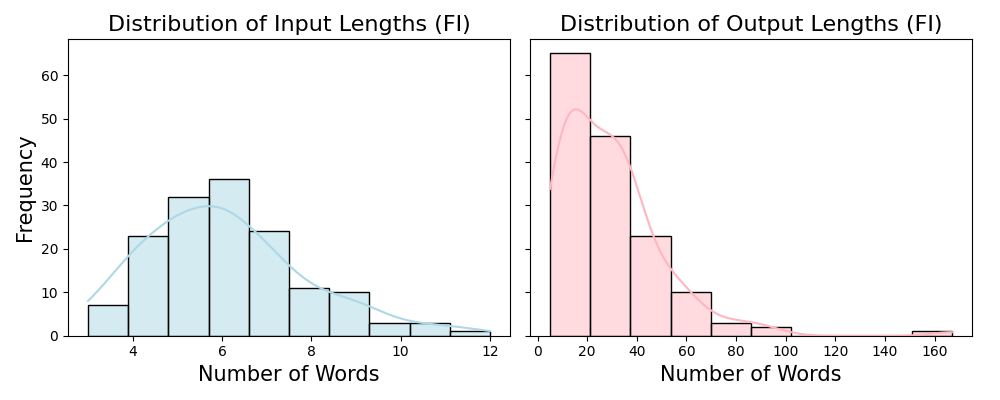}
    \caption{Input/Output length distribution for Finnish.}
    \label{fig:enter-label}
\end{figure}

\begin{figure}[h!]
    \centering
    \includegraphics[width=0.99\linewidth]{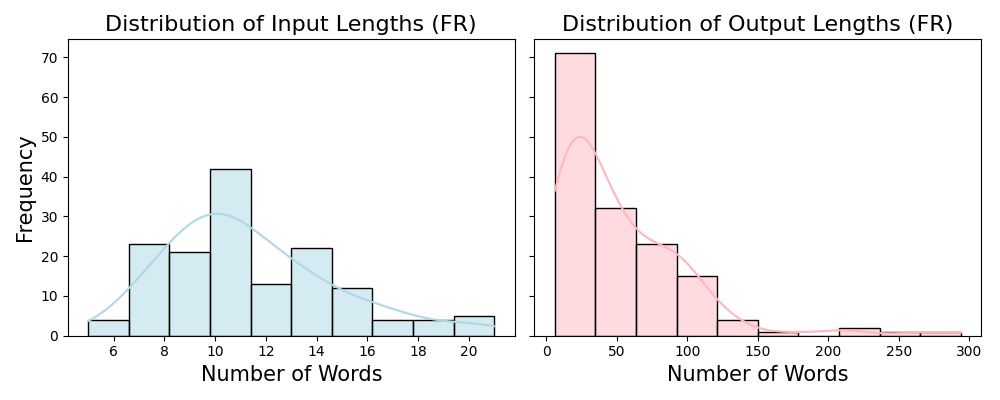}
    \caption{Input/Output length distribution for French.}
    \label{fig:enter-label}
\end{figure}

\begin{figure}[h!]
    \centering
    \includegraphics[width=0.99\linewidth]{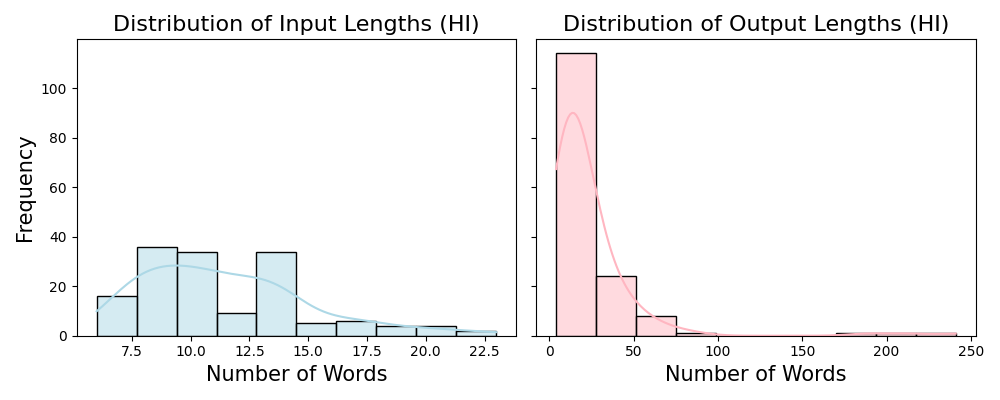}
    \caption{Input/Output length distribution for Hindi.}
    \label{fig:enter-label}
\end{figure}

\begin{figure}[h!]
    \centering
    \includegraphics[width=0.99\linewidth]{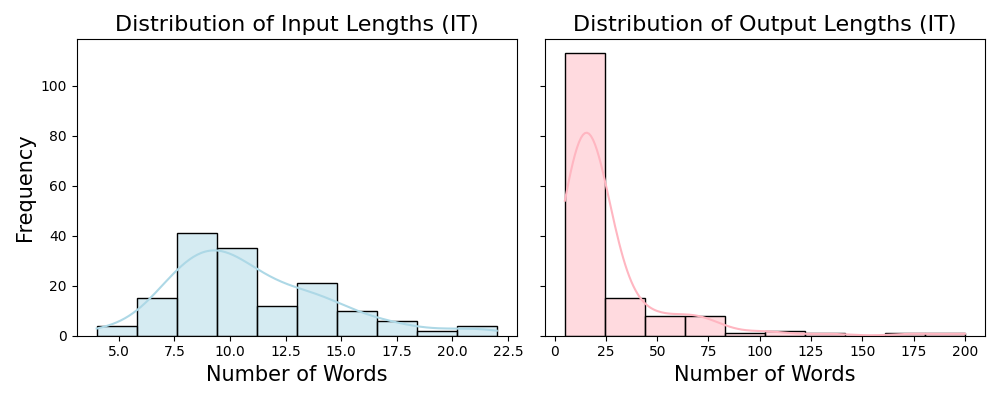}
    \caption{Input/Output length distribution for Italian.}
    \label{fig:enter-label}
\end{figure}

\begin{figure}[h!]
    \centering
    \includegraphics[width=0.99\linewidth]{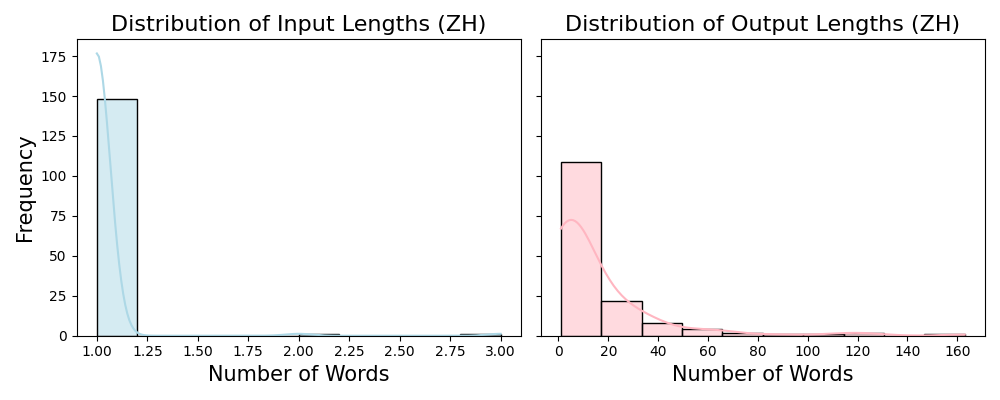}
    \caption{Input/Output length distribution for Chinese.}
    \label{fig:enter-label}
\end{figure} 
\begin{figure*}[h!]
    \centering
    \includegraphics[width=0.62\linewidth]{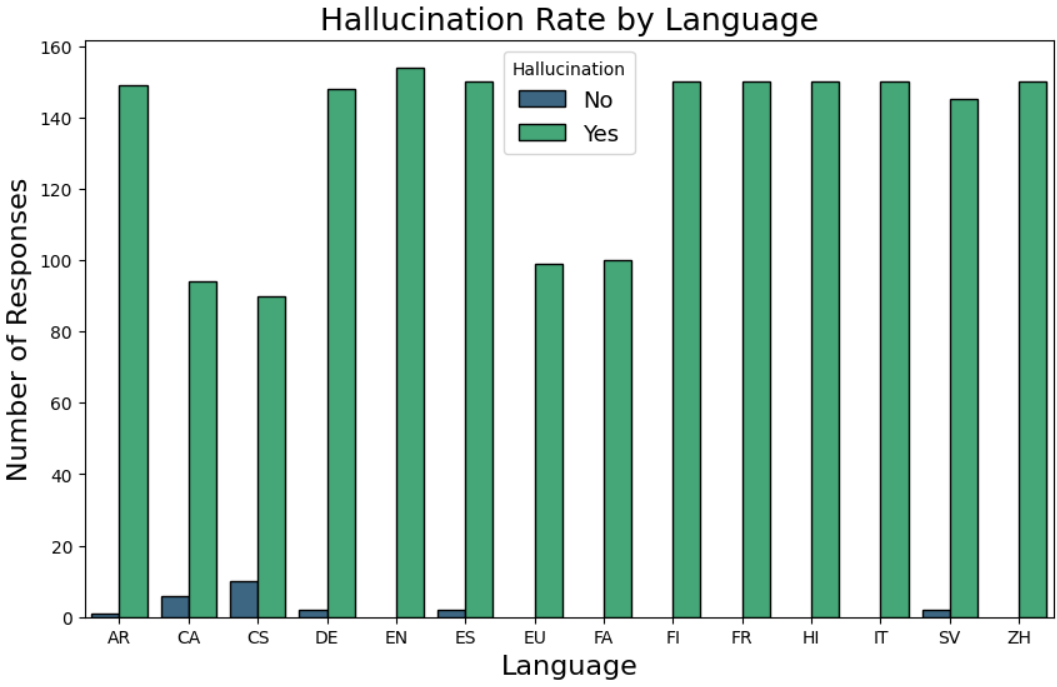}
    \caption{Hallucination rate per language according to 'soft label' annotations.}
    \label{fig:hal-rate}
\end{figure*}
\section{Examples}
We present some input/output examples from respective LLMs in Table \ref{tab:languages-examples}.
\begin{table*}[h]
    \centering \small
    \begin{tabular}{c|p{4cm}p{9.7cm}}
\hline
Lang & Input & Output \\
\hline
EU&Zenbat Pentastomida espezie daude?&Hiru espezie bakarrik daude.\\
\hline
DE&Welche Filme produzierte David Sandberg im Jahr 2014?&David Sandbergs Filme im Jahre 2014 waren "The Amazing Spider-Man 2: Rise of Electro" (2014), "Die Tribute von Panem - Mockingjay Teil 2" ("The Hunger Games: MOCKINGJAY - Part 2", 2014), und "Der Hobbit: Die Schlacht der fünf Heere" "(The Hobit: The Battle of the Five Armies", 2015).\\
\hline
FR&À la suite de quel grand événement a été créee la ville de La Bellière ?&La ville de La Bellière a été créée en 1975 après la fusion de cinq communes voisines : Bellière, Bois-Jérôme, Bretteville-l'Orgueilleuse, Saint-Martin-d'Écublei et Saint-Michel-de-la-Pierre. Le motif de la fusion était la volonté des communes de coopérer pour améliorer leurs infrastructures et offrir un meilleur niveau de services aux résidents. Ce n'y a pas d'événement particulier qui a déclenché la création de la ville, mais plutôt une décision concertée des conseils municipaux de ces cinq communes pour s'unir pour des raisons économiques et administratives.\\\hline
ZH& \begin{CJK}{UTF8}{gbsn}大卫·桑德柏格的导演处女作是什么？\end{CJK}&
\begin{CJK}{UTF8}{gbsn}大卫桑德的导演处子秀是1993年的电影《性、谎言和录像带》。这部电影获得了戛纳电影节金棕榈奖，并成为当时独立电影的里程碑之一。\end{CJK}\\\hline

IT&Da chi fu coniato il termine "paradosso francese"?&Il termine "paradosso francese" fu coniato dal nutrizionista americano Ancel Keys nel 1952.\\\hline
CA&A la vora de quin mar es troba la ciutat russa Kaspíisk?&Kaspíisk es una ciutat situada a la costa del mar de Azov, a la província de Krasnodar, a la Rússia.\\\hline

ES&¿En qué equipos de la NHL jugó Bert Olmstead durante su carrera?&Bert Olmstead jugó para los Montreal Canadiens, Toronto Maple Leafs, New York Rangers y Los Angeles Kings durante su carrera en la NHL.\\\hline

FI&Millä nimellä David Sandberg tunnetaan YouTubessa?&David Sandberg tunnetaan paremmin nimellä "AtomicDave" YouTubessa. Hän on tunnettu sisällöntuottaja, joka julkaisee usein videoita, joissa hän pelaa pelejä ja tekee haasteita. Hänellä on yli miljoona tilaajaa YouTubessa, ja hän on tunnettu energisestä ja viihdyttävästä tyylistään.\\\hline

EN&Did Alberto Fouillioux ever play in a world cup championship?& No, Albero Foulois was not in any of the FIFA World Cup finals.\\\hline

SV&I vilken stad föddes regissören David Sandberg?&David Sandbergs födelseort är New York.\\\hline

CS&
\foreignlanguage{czech}{Jaká oblast České republiky je označována slovem sudety?}&
\foreignlanguage{czech}{Sudety je označení pro horské pásmo v České republice a Polsku, které se táhne od severozápadu na jihovýchod. V České republice se nacházejí Sudetské hory, které se dělí na Krkonoše, Jeseníky, Orlické hory, Adršpach-Teplické skály, Rychlebské hory, Hrubý Jeseník, Javoří hory, Zlaté hory a Králický Sněžník. Tyto hory byly důležitým místem během druhé světové války, kdy se zde odehrávaly bitvy mezi německými a československými vojsky.}\\
\hline
    \end{tabular}
    \caption{Examples (inputs/LLM outputs) in MuSHROOM languages.}
    \label{tab:languages-examples}
\end{table*}

\section{Preliminary experiments}
\label{sec:preliminaries}
In order to deploy our LLM-based system, we conduct some exploratory experiments with Llama and Claude. For that purpose we initially employ the labeled test set from the SHROOM-shared task of 2024 \cite{shroom-2024-semeval}, due to its larger amount of labeled examples. This dataset contains instances with the following features: \textit{Source - src} is the input given to a model, \textit{hypothesis - hyp} is the output generated by the model, \textit{target - tgt} comprises the ground truth output for this specific model, \textit{reference - ref}
indicates whether target, source or both of these fields contain the semantic information necessary to establish whether a datapoint is a hallucination, \textit{task} refers to the task being solved and \textit{model} to the model being used (in the model-agnostic case
the model entry remains empty).

In this task, the participants were asked to classify the output of the LLM as hallucination or not based on the meaning of the target output that the LLM should have produced. Each instance also contains the tag \textit{Hallucination/Not Hallucination} and a \textit{probability} expressing the ratio of the annotators that marked the output as Hallucination over all the annotator that participated.

In order to explore the capabilities of Llama and Claude in hallucination detection, we manipulate the SHROOM-2024 dataset by bringing it to the format of this year’s dataset. To perform that we create the feature 'model input' by combining the source and the task feature that has three distinct values (MT - Machine Translation, PG - Paraphrase Generation, D - Definition Modeling) as described in the Table \ref{tab:model-input-feat}.

\begin{table*}[h!]
    \centering \small
    \begin{tabular}{c|p{1.2cm}|p{6.5cm}}
\hline
Task & Source & Model Input \\
\hline
Machine Translation (MT) & src & Translate the following sentence in English : \{src\} \\
Definition Modelling (DM) & src & \{src\} \\
Paraphrase Generation (PG) & src & Paraphrase the following sentence : \{src\} \\
\hline
    \end{tabular}
    \caption{Combination of task identifier and source text to create model input feature}
    \label{tab:model-input-feat}
\end{table*}

\paragraph{Preliminary experiments} involve probing the hallucination detection capabilities of Llama and Claude models. More specifically we conduct the following four experiments:

\textbf{1. Zero-shot (ZS)} We use a simple zero-shot prompt that questions in the form: \textit{''Given the \{input\} and the \{output\}, is the output a hallucination? Reply with Yes/No''}. 

\textbf{2. ZS + Hypothesis} To further boost this, we also leverage the \textit{hypothesis} provided, transforming the prompt as: \textit{''Given the {input}, the {output} and the {hypothesis}, is the output a hallucination? Reply with Yes/No''}.

\textbf{3. CoT + Hypothesis} Since the results' accuracy was close to randomly assigning a Hallucination/not Hallucination tag, instead of prompting the LLMs to simply respond with a binary Yes/No label, we prompt them to develop their thoughts; then, based on this and the hypothesis, we generate the final Yes/No label. The prompt used in this case is: \textit{"Given the \{input\}, \{output\} and the \{hypothesis\} is the output a hallucination? Firstly, explain your thought and in the end write the word Yes or No if it is or it is not a hallucination respectively."}

\begin{table*}[h!]
    \centering \small
    \begin{tabular}{l|c|c|c|c}
\hline
Model & ZS & ZS + Hypothesis & CoT + Hypothesis & Hypothesis Generation \\
\hline
Llama & 0.52 & 0.55 & 0.62 & 0.79\\
Claude & 0.51 & 0.55 & 0.63 & 0.83\\
\hline
    \end{tabular}
    \caption{Results of preliminary experiments: Columns ZS, ZS + Hypothesis, CoT + Hypothesis refer to the accuracy of the classification task (Hallucination/Not Hallucination) and the Hypothesis Generation refer to the similarity of the produced answers with the ground truth provided}
    \label{tab:preliminary-exp1}
\end{table*}

\begin{table*}[h]
    \centering \small

    \begin{tabular}{c|c|c|c|c|c|c}

\hline
    Metric & \textbf{C}&
    \textbf{L} & \textbf{C, CH} & \textbf{C, LH} & \textbf{L, LH} &  \textbf{L, CH}\\
\hline
    IoU& 0.465 & 0.477&0.478&0.521&0.491&0.543\\
    Cor&0.525 &0.433&0.587&0.525&0.522&0.582\\
    
\hline
    \end{tabular}
    \caption{ IoU and Correlation scores for the english: C: Claude, No Hypothesis, L: Llama, No Hypothesis, C,CH: Claude + Claude Hypothesis, C,LH: Claude + Llama Hypothesis , L, LH: Llama +Llama Hypothesis , L,CH: Llama + Claude Hypothesis}
    \label{tab:preliminary-ex-EN}
\end{table*}

\begin{table*}[h!]
    \centering \small
    \begin{tabular}{l|c|l|c|l|cc}

\hline
     &Metrics& &Metrics&&Metrics&\\
\hline
    \textbf{C+L}&IoU:0.42 Cor: 0.57 &\textbf{L + C,LH}&IoU: \textbf{0.52 Cor: 0.61}&\textbf{L, LH +L, CH}&IoU: 0.41 Cor: 0.57\\
    \textbf{C+ C,CH}&IoU: 0.50 Cor: 0.57 &\textbf{L + L, LH}&IoU: 0.39 Cor: 0.51&\textbf{C, CH+ L,LH}&IoU: 0.46 Cor: 0.61  \\
    \textbf{C + C, LH}&\textbf{IoU: 0.53 Cor: 0.59}&\textbf{L + L, CH}&IoU: 0.51 Cor: 0.62&\textbf{C, CH + L, CH}&IoU: 0.46 Cor: 0.55\\
\textbf{C + L, LH}&IoU: 0.44 Cor: 0.59&\textbf{L+ C, CH}&IoU: 0.44
Cor: 0.59 &\textbf{C, LH + L, LH}&IoU: 0.44 Cor: 0.612\\
\textbf{C +L, CH}&\textbf{IoU: 0.53 Cor: 0.59} &\textbf{C, CH +C, LH}&IoU: 0.50 Cor: 0.62&\textbf{C, LH +L, CH}&\textbf{IoU: 0.54 Cor: 0.64}\\
    
\hline
    \end{tabular}
    \caption{IoU and Correlation scores for the combination of the results from different experiments: C: Claude, No Hypothesis, L: Llama, No Hypothesis, C,CH: Claude + Claude Hypothesis, C,LH: Claude + Llama Hypothesis , L, LH: Llama +Llama Hypothesis , L,CH: Llama + Claude Hypothesis}
    \label{tab:preliminary-ex2-EN}
\end{table*}
\textbf{4. Hypothesis generation and similarity} Since the Mu-SHROOM dataset does not contain ground truth hypotheses, we prompt the LLMs to generate those by replying to the input query, so that we can assess the hypothesis similarity with the generated output using Natural Language Inference (NLI) models, similar to \citet{grigoriadou-etal-2024-ails}. 
The results are presented in Table \ref{tab:preliminary-exp1}.

After these experiments, we manually observe the false negatives, and conclude the following:

1. Even though Claude performs better than Llama in general, there are cases where Llama is able to detect some hallucinations that Claude could not.

2. Even though providing a hypothesis improves the results, there are cases that the LLM (either of the two) focuses more on the hypothesis than its internal knowledge and therefore it fails to recognize a hallucinated part.

On the second part of the preliminary experiments, we tried combining different components in order to benefit from the extra information that the combinations provide and then choose the best strategy for the development of our final system. Those experiments were conducted with the validation set in order to gain some insight on the performance of the models with the multilingual texts and are the following:

\textbf{1. No Hypothesis} In the first experiment we prompt Claude and Llama to detect hallucination spans without providing a hypothesis.

\textbf{2. Model + Hypothesis from the same model} In the second experiment we prompt Claude and Llama to generate answers for the input texts, and then prompt them to detect hallucination spans on the output text given the answers each model produced, i.e. prompt Llama given as hypothesis the answers that Llama produced and Claude given as hypothesis the answers that Claude produced.

\textbf{3. Model + Hypothesis from the other Model} In the third experiment we prompt Claude and Llama to generate answers for the input texts, and then prompt them to detect hallucination spans on the output text given the answers the other model produced, i.e. prompt Llama given as hypothesis the answers that Claude produced and Claude given as hypothesis the answers that Llama produced.

\textbf{4. Combinations of two of the above} For the fourth experiment we examine how the combinations of the results of two of the components above improve the performance.

To measure these results we used the evaluation metrics of the task (IoU and Cor) but we prioritized the IoU to choose the dominant components.

In Tables \ref{tab:preliminary-ex-EN} , \ref{tab:preliminary-ex2-EN} we present the results for each experiment in english in detail and then we present the results for the top-3 strategies for every other language.

\begin{table}[t!]
    \centering \small
    \renewcommand{\arraystretch}{0.6}
    \setlength{\tabcolsep}{4.2pt}
    
    \begin{tabular}{c|l}
    \hline
    \textbf{Language id} &  \textbf{Experiment, IoU}\\
    \hline
    
    \multirow{3}{5em}{\textbf{ar}} & Claude, Llama Hypothesis: 0.53\\
       & Llama, Claude Hypothesis: 0.52\\
        &Claude, no Hypothesis: 0.49
       \\
    
    \hline
    \multirow{3}{5em}{\textbf{de}}& Llama, Claude Hypothesis: 0.56\\
        &Claude, Llama Hypothesis: 0.55\\
        &Llama, no Hypothesis: 0.54
     \\
     \hline
    
     \multirow{3}{5em}{\textbf{es}}& Llama, Claude Hypothesis: 0.42\\
        &Llama, no Hypothesis: 0.41\\
        &Claude, Llama Hypothesis: 0.49
       \\
       \hline
    
    \multirow{3}{5em}{\textbf{fi}}& Llama, Claude Hypothesis: 0.61\\
      &Claude, Llama Hypothesis: 0.59\\
      &Llama, no Hypothesis: 0.56
    \\
    \hline

    \multirow{3}{5em}{\textbf{fr}}& Claude, Claude Hypothesis: 0.58\\
     & Claude, Llama Hypothesis: 0.57\\
      &Claude, no Hypothesis: 0.56
       \\
       \hline
    
    \multirow{3}{5em}{\textbf{hi}}& Llama, Claude Hypothesis: 0.57\\
      &Claude, Llama Hypothesis: 0.55\\
      &Llama, no Hypothesis: 0.52
      \\
      \hline
    
    \multirow{3}{5em}{\textbf{it}}& Llama, Claude Hypothesis: 0.60\\
        &Claude, Llama Hypothesis: 0.59\\
        &Llama, Llama Hypothesis: 0.59
       \\
       \hline
    
   \multirow{3}{5em}{\textbf{sv}}& Claude, Llama Hypothesis: 0.51\\
        &Llama, Claude Hypothesis: 0.47\\
        &Llama, Llama Hypothesis: 0.44
      \\
      \hline
    
    \multirow{3}{5em}{\textbf{zh}}& Claude, Llama Hypothesis: 0.35\\
        &Claude, Claude Hypothesis: 0.33\\
        &Llama, Llama Hypothesis: 0.30
       \\
       \hline

    \hline
    \end{tabular}
    \caption{Best Components for each language}
    \label{tab:preliminary_exp2}
\end{table}

\begin{table*}[ht!]
    \centering \small

    \begin{tabular}{l|c|c|c|c}

\hline
    \textbf{Components} & \textbf{IoU1}&\textbf{IoU2}&\textbf{IoU3}&
    \textbf{IoU4} \\
\hline
    Claude, no Hypothesis + Claude, Llama Hypothesis&0.47&0.52& 0.465 & \textbf{0.53}\\
    Claude, Llama Hypothesis + Claude, Claude Hypothesis&0.52&0.48&0.87 &0.50\\
    
\hline
    \end{tabular}
    \caption{For combinations of components, we calculate the IoU of each separate component with the reference values (IoU1, IoU2), the IoU between the predictions of different components (IoU3) and the IoU between the predictions of the combined components and the reference values (IoU4)}
    \label{tab:ious}
\end{table*}

For the last experiment, we calculated the IoU of the predictions that occurred from the different experiments and we came to the conclusion that even though the IoU between predictions with and without hypothesis are lower than these of prediction with hypothesis from different sources, the combination of predictions with and without hypothesis reached better scores. After carefully inspecting the results, we found that reasonable because the lack of hypothesis led the LLM to emphasize on more parts that might contain hallucinations rather than the factual inconsistencies that were more often inspected when the hypothesis was provided. An example is provided for English in Table \ref{tab:ious}.

Based on these findings showing how the LLMs benefit from the answers provided by the other LLM, we design the prompting experiments presented in the main paper (Section \ref{sec:overview} - Prompting strategies).

\section{Prompts}
\label{sec:prompts}
The prompts to initialize the hallucination detection and the answer generation processes are presented in Tables \ref{tab:initprompts}, \ref{tab:initprompts2}. The system and user prompts designed for our approaches are presented in Tables \ref{tab:example}, \ref{tab:example-2}.
\begin{table*}[h!]
    \centering\small

    \begin{tabular}{p{3cm}|p{13cm}}
        \hline
    \textbf{Description} & \textbf{Prompt} \\
     \hline
    \multicolumn{2}{c}{\textbf{Zero-Shot Scenario}}\\
     \hline

    \textit{Hallucination Definition}& You are a hallucination detector. A hallucination is the production of fluent but incorrect output of an LLM.The definition of incorrect output falls in four categories:     

a. The output is inconsistent with the input, so the produced answer does not answer the input query or is irrelevant to it.,

b. The output contains a factual inconsistency, so contains something that is not a validated fact or is wrong.
                
c. The output contains contradictory facts so in the output there are things that cannot be true at the same time.
                 
d. The output contains mispelled words\\
\hline
\textit{Output Format Instruction}& I will provide some examples for you to find hallucinations based on the given definition of hallucination. Although there might be several parts where hallucinations of probably different types occur, the answer should only end with the phrase 'So the hallucinations are: ' followed by the hallucinations exactly as they are written in the sentence given, inside "" and separated by commas  \\ 
   \hline
   \multicolumn{2}{c}{\textbf{Few-Shot Scenario}}\\
     \hline
    \multirow{3}{5em}{\textit{One example for each hallucination type}}&\textbf{Example 1(input-output conflict)}:The input is: Where did the Olympic Games of 2004 take place? The output is: The Olympic Games of 2020 took place in London.As a hallucination detector you should point out that there is a hallucination here because the output replies the answer where did the Olympic Games of 2020 take place. So the hallucinations are:"2020".\\
               &
    \textbf{Example 2 (factual inconsistency)}:The input is: Where did the Olympic Games of 2004 take place? The output is: The Olympic Games of 2004 took place in Florida.As a hallucination detector you should point out that there is a hallucination here because the Olympic Games of 2004 took place in Athens, so there is a factual inconsistency in the word "Florida".So the hallucinations are: "Florida"\\
& 
\textbf{Example 3(internal output conflict)}:The input is: Where did the Olympic Games of 2004 take place and what was the biggest Stadium used? The output is: The Olympic Games of 2004 took place in Athens, Greece. All stadiums were designed for that purpose but the biggest was Olympic Stadium of Athens "Spyros Louis" that was built in 1982.As a hallucination detector you should point out that there is a hallucination here because the output states that all stadiums were created for that purpose but the Olympic Stadium of Athens "Spyros Louis" was built in 1982 so much earlier than the Olympic Games, So the hallucinations are: "All stadiums were designed for that purpose". \\
\hline
   \end{tabular}
\caption{Prompts used to initialize the hallucination detection or the answer generation process.}
    \label{tab:initprompts}
\end{table*}

\clearpage

\begin{table*}[h]
    \centering\small

    \begin{tabular}{p{4cm}|p{9cm}}
        \hline
    \textbf{Description} & \textbf{Prompt} \\

    \hline
    \multicolumn{2}{c}{\textbf{Few-Shot Scenario}}\\
     \hline
    
\textit{One example for each hallucination type}&\textbf{Example 4(misspelling)}:The input is: Where did the Olympic Games of 2004 take place? The output is: The OLympooooc Games of 2004 took place in Athens, Greece.
                
As a hallucination detector you should point out that there is a hallucination here because the OLympoooooc Games are a misspelling of the Olympic Games So the hallucinations are: "OLympooooc".\\

\hline
\textit{One example for the expected output format}
 & \textbf{Example(output format)}  The input is: What is the biggest church in Greece? The output is: The biggest church in Greece is Saint George located in the center of Athens. It has a maximum length of 73 m and width 48 m and it is the biggest church of Greece.The church is in the downtown of the modern city of Athens, close to the high-traffic Acharnon Avenue.The foundations of the church were laid on 12 September 1910 by King George I of Greece and it was consecrated on 10 JUly 1935.

The expected output is: There are several hallucinations. The name of the biggest church is Saint Panteleimon of Acharnai and is indeed located in the center of Athens but has a maximum length of 63 m and was consecrated on 22 June 1930. So the hallucinations are: "saint George","73","10 JUly 1935".\\ 
   
   \hline
   \multicolumn{2}{c}{\textbf{Answer generation}}\\
   \hline
   \textit{Answer Generation}& I will provide a question and you will provide the answer
   \\
\hline
   \end{tabular}
\caption{Continuation of table \ref{tab:initprompts}. Prompts used to initialize the hallucination detection or the answer generation process.}
    \label{tab:initprompts2}
\end{table*}

\begin{table*}[h]
    \centering\small

    \begin{tabular}{p{3cm}|p{9cm}|p{3cm}}
        \hline
    \textbf{Approach} & \textbf{System Prompt} & \textbf{User Prompt} \\
     \hline
    \textbf{Preliminary Test}: input-output pair +specific part of the output to assign a Hallucination/not Hallucination tag & You are a hallucination detector. I will provide some input-output pairs and a specific part of the output and you have to decide whether the specific part is a hallucination as it was defined, based on your sources of knowledge. Write 'Hallucination' or 'Not Hallucination' if it is a hallucination or not respectively. The tag 'Hallucination' or 'Not Hallucination' should be the only words in your answer. & The input is:\textit{input}, the output is :  \textit{output} and the part that might contain hallucination is: \textit{part}

   \\ 
   
    \hline
    \textbf{Input-output pair} & You are a hallucination detector. I will provide some input-output pairs that represent inputs given to LLMs and the outputs they produced.Define the specific words or parts of the outputs that are hallucinations based on your sources of knowledge.Try to specify as much as possible the parts that are a hallucination even if it is just one word and put those parts in "".Inside "" include only parts in the exact way they are written in the given sentence.In your answer explain your thought and then provide the parts seperated with commas in the end of your answer after the sentence 'So the hallucinations are:.After this sentence include only parts of the output in the exact way they are written and nothing more & The input is:\textit{input} and the output that might contain hallucination is: \textit{output}

   \\ 
   
   \hline
   \textbf{Input-output pair + \newline Hypothesis} & You are a hallucination detector. I will provide some input-output pairs that represent inputs given to LLMs and the outputs they produced and a hypothesis. Define the specific words or parts of the outputs that are hallucinations based on your sources of knowledge and the hypothesis.Try to specify as much as possible the parts that are a hallucination even if it is just one word and put those parts in "".Inside "" include only parts in the exact way they are written in the given sentence. In your answer explain your thought and then provide the parts seperated with commas in the end of your answer after the sentence "So the hallucinations are:" After this sentence include only parts of the output in the exact way they are written and nothing more & The input is:\textit{input},  the output that might contain hallucination is: \textit{output} and the hypothesis is \textit{hypothesis}
   \\
\hline
 
    \end{tabular}
\caption{Prompts (system and user) regarding our various prompting and translation approaches.}
    \label{tab:example}
\end{table*}

\clearpage

\begin{table*}[h]
    \centering\small

    \begin{tabular}{p{3cm}|p{9cm}|p{3cm}}
        \hline
    \textbf{Approach} & \textbf{System Prompt} & \textbf{User Prompt} \\ 
\hline
 \textbf{Input-output pair + \newline English Translations}& You are a hallucination detector. I will provide some input-output pairs that represent inputs given to LLMs and the outputs they produced.
   The task is to define the specific words or parts of the outputs that are hallucinations based on your sources of knowledge.
   
   The given input-output pairs are in different languages.If they are not in english I will also provide a translation in English. The process you should follow includes some steps:
    
    1. Examine the translation if provided or the original sentence if it is in english
    
    2. Try to specify as much as possible the parts that are a hallucination even if it is just one word.
    
    3. Explain your thoughts in English and then put the parts of the original sentence in the original language that correspond to the hallucinated parts you detected in English in "". Inside "" include only parts in the exact way they are written in the original sentence.
    So in your answer explain your thought in english and then provide the parts in the original language, separated with commas in the end of your answer after the sentence 'So the hallucinations are:.
    After this sentence include only parts of the output in the exact way they are written and nothing more. & The input is:\textit{input} and the output that might contain hallucination is: \textit{output}. The english translation of the input is \textit{input translation} and the english translation of the output is \textit{output translation}.
    \\
\hline
   \textbf{Input-output pair + \newline English Translations + \newline Hypothesis} & You are a hallucination detector. I will provide some input-output pairs that represent inputs given to LLMs and the outputs they produced.
    The task is to define the specific words or parts of the outputs that are hallucinations based on your sources of knowledge and a provided hypothesis.
    The given input-output pairs are in different languages.If they are not in english I will also provide a translation in English. The process you should follow includes some steps:
    
    1. Examine the translation if provided or the original sentence if it is in english.
    
    2. Try to specify as much as possible the parts that are a hallucination even if it is just one word.
    
    3. Explain your thought in english and then put the parts of the original sentence in the original language that correspond to the hallucinated parts you detected in english in "".
    
    Inside "" include only parts in the exact way they are written in the original sentence.
    So in your answer explain your thought in english and then provide the parts in the original language, seperated with commas in the end of your answer after the sentence 'So the hallucinations are:.
     After this sentence include only parts of the output in the exact way they are written and nothing more. 
 & The input is:\textit{input} and the output that might contain hallucination is: \textit{output}. The english translation of the input is \textit{input translation} and the english translation of the output is \textit{output translation} and the hypothesis is \textit{hypothesis}.
 \\
\hline
   \textbf{Input-output pairs + \newline LLM Translation}  & You are a hallucination detector. I will provide some input-output pairs that represent inputs given to LLMs and the outputs they produced.
The task is to define the specific words or parts of the outputs that are hallucinations based on your sources of knowledge.
The given input-output pairs are in different languages. So the process you should follow includes some steps:

1. Translate the input and output in English

2. Try to specify as much as possible the parts that are a hallucination even if it is just one word

3. Explain your thought in english and then put the parts of the original sentence in the original language that correspond to the hallucinated parts you detected in english in "".
Inside "" include only parts in the exact way they are written in the given sentence.
So in your answer explain your thought in english and then provide the parts in the original language, separated with commas in the end of your answer after the sentence 'So the hallucinations are:’.After this sentence include only parts of the output in the exact way they are written and nothing more. If the given input-output pairs are in english do not do the translation step.'
& The input is:\textit{input} and the output that might contain hallucination is: \textit{output}. \\
 \hline
    \end{tabular}
\caption{Continuation of Table \ref{tab:example}. Prompts (system and user) regarding our various prompting and translation approaches.}
    \label{tab:example-2}
\end{table*}

\end{document}